\documentclass[conference]{IEEEtran}
\IEEEoverridecommandlockouts
\usepackage{cite}
\usepackage{amsmath,amssymb,amsfonts}
\usepackage{algorithmic}
\usepackage{graphicx}
\usepackage{textcomp}
\usepackage{float}
\usepackage{caption}
\usepackage{listings}
\usepackage{booktabs} 
\usepackage{multirow} 
\usepackage{xcolor}
\def\BibTeX{{\rm B\kern-.05em{\sc i\kern-.025em b}\kern-.08em
    T\kern-.1667em\lower.7ex\hbox{E}\kern-.125emX}}
\begin{document}

\title{Evaluating Multimodal Large Language Models on Educational Textbook Question Answering\\
%{\footnotesize \textsuperscript{*}Note: Sub-titles are not captured for https://ieeexplore.ieee.org  and should not be used}
\thanks{
© 2025 IEEE. Personal use of this material is permitted. 
Permission from IEEE must be obtained for all other uses, in any current or future media, including reprinting/republishing this material for advertising or promotional purposes, 
creating new collective works, for resale or redistribution to servers or lists, or reuse of any copyrighted component of this work in other works.

This work has been accepted to the 2nd International Generative AI and Computational Language Modelling Conference (GACLM 2025) for publication in the proceedings.
}
}

\author{
\centering
\begin{tabular}{c c}
% -------- Row 1 --------
\begin{minipage}[t]{0.45\textwidth}
\centering
\textbf{1\textsuperscript{st} Hessa A. Alawwad} \\
\small Faculty of Computing and Information Technology \\
\small King Abdulaziz University \\
\small Jeddah, Saudi Arabia \\
\small alawwad.a.hessa@gmail.com
\end{minipage}
&
\begin{minipage}[t]{0.45\textwidth}
\centering
\textbf{2\textsuperscript{nd} Anas Zafar} \\
\small FAST School of Computing \\
\small National University of Computer and Emerging Sciences \\
\small Karachi, Pakistan
\end{minipage}
\\
\multicolumn{2}{c}{\vspace{2ex}} \\  % <<< actual space between Row 1 and 2

% -------- Row 2 --------
\begin{minipage}[t]{0.45\textwidth}
\centering
\textbf{3\textsuperscript{rd} Areej Alhothali} \\
\small Faculty of Computing and Information Technology \\
\small King Abdulaziz University \\
\small Jeddah, Saudi Arabia
\end{minipage}
&
\begin{minipage}[t]{0.45\textwidth}
\centering
\textbf{4\textsuperscript{th} Usman Naseem} \\
\small School of Computing \\
\small Macquarie University \\
\small Sydney, Australia
\end{minipage}
\\
\multicolumn{2}{c}{\vspace{2ex}} \\  % <<< actual space between Row 2 and 3

% -------- Row 3 --------
\begin{minipage}[t]{0.45\textwidth}
\centering
\textbf{5\textsuperscript{th} Ali Alkhathlan} \\
\small Faculty of Computing and Information Technology \\
\small King Abdulaziz University \\
\small Jeddah, Saudi Arabia
\end{minipage}
&
\begin{minipage}[t]{0.45\textwidth}
\centering
\textbf{6\textsuperscript{th} Amani Jamal} \\
\small Faculty of Computing and Information Technology \& \\
\small Center of Research Excellence in AI and Data Science \\
\small King Abdulaziz University \\
\small Jeddah, Saudi Arabia
\end{minipage}
\end{tabular}
}

\maketitle

\begin{abstract}
Multimodal large language models (MLLMs) have shown success in vision-language tasks, but their ability to reason over complex educational materials remains largely untested. This work presents the first evaluation of state-of-the-art MLLMs, including LLaVA-1.5 and LLaMA 3.2-Vision, on the textbook question answering (TQA) task using the CK12-QA dataset. We introduce a multimodal retrieval-augmented generation (RAG) pipeline to simulate real-world learning by providing relevant lesson paragraphs and diagrams as context. Our zero-shot experiments reveal a critical trade-off; while retrieved context improves LLaVA's performance on text-based questions, it significantly degrades the accuracy of the more powerful LLaMA 3.2-Vision on diagram-based tasks, dropping its validation accuracy from 74.07\% to 25.93\%. We term this statistically significant phenomenon "catastrophic context interference." Furthermore, fine-tuning highlights architectural differences; LLaMA 3.2-Vision’s performance substantially improves to 71.16\% on the test set, demonstrating its capacity to learn multimodal integration, whereas LLaVA's performance declines, indicating challenges with generalization. Our results underscore the challenges MLLMs face in modality prioritization and context integration, providing a benchmark and pointing to key directions for developing more robust AI-driven educational tools.
\end{abstract}


\section{Introduction}
\label{sec:introduction}
Answering curriculum-related questions in multimodal educational materials is a central challenge in AI for education, requiring systems to reason across complex multimodal contexts such as lengthy lessons, diagrams, and videos. While general-purpose visual question answering (VQA) has made substantial progress with multimodal large language models (MLLMs) such as LLaVA~\cite{liu2023llava, liu2023improvedllava} and LLaMA 3.2-Vision~\cite{grattafiori2024llama}, most evaluations have focused on natural images or open-domain reasoning, overlooking the distinct demands of textbook question answering (TQA). TQA requires not only parsing long-form lessons but also interpreting intricate diagrams, integrating supplementary resources such as videos, and analyzing comprehensive, often fragmented, curricular content that exceeds the scope of standard VQA tasks~\cite{bewersdorff2025taking}. This underscores the significance of our contribution in addressing the challenges of TQA.

In science and education, where students must comprehend information from diverse sources such as plain language, diagrams, and tables, the need for systems capable of multimodal learning and understanding becomes especially critical. The integration of multimodal large language models (MLLMs) into educational contexts offers promising opportunities for individualized support, adaptive learning, and deeper assessment. However, it also raises significant concerns related to bias, data alignment, accuracy, and the ongoing necessity for human oversight~\cite{kuchemann2025opportunities}. 

To address these gaps, our work focuses on the CK12-QA dataset~\cite{kembhavi2017you}, a comprehensive benchmark comprising over 26,000 questions derived from the CK-12 Foundation’s open educational resources. The dataset includes both text-only (NDQ) and diagram-based questions (NDQ), each paired with corresponding lesson texts and images to simulate realistic curriculum-based reasoning scenarios. By augmenting the questions with retrieved lesson content, we develop a retrieval-augmented generation (RAG) pipeline that more closely reflects real-world study practices and enables a rigorous evaluation of MLLMs’ ability to integrate and reason across multimodal contexts.

\begin{figure}[tb]
\centering
\includegraphics[width=\linewidth]{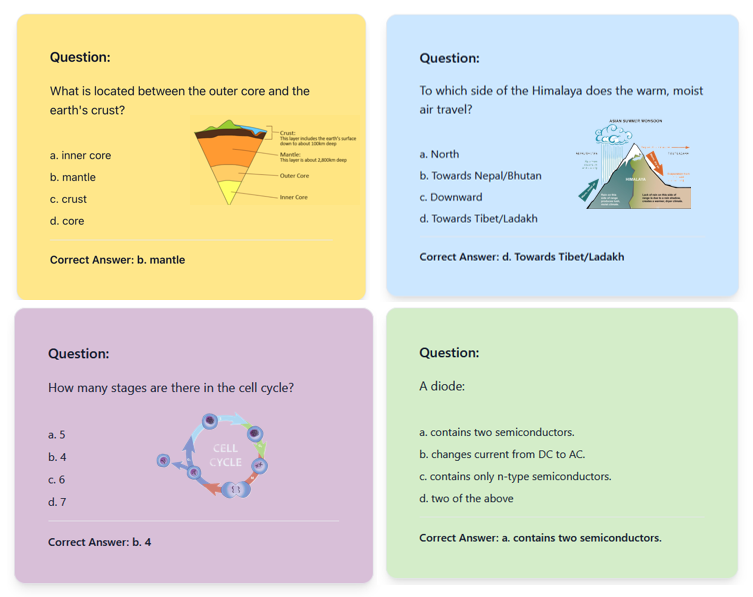}
\vspace{6pt}
\captionsetup{labelfont=bf, labelsep=period, justification=raggedright}
\caption{Examples from the CK12-QA dataset~\cite{kembhavi2017you}, illustrating various question types, including both diagram-based and non-diagram-based formats.}
\label{fig:dataset_examples}
\end{figure}

The CK12-QA questions illustrated in Fig.~\ref{fig:dataset_examples} are multimodal, multiple-choice science questions designed  

\noindent to engage students in mid-level cognitive tasks, including interpretation, conceptual reasoning, and visual comprehension. Unlike simple fact-recall questions, these items require a solid understanding of scientific concepts, such as the layers of the Earth, airflow over mountains, stages of the cell cycle, and semiconductor behavior, and the ability to synthesize textual prompts with informative diagrams. For example, students must deduce correct answers by analyzing labeled images or interpreting scientific processes. This aligns the questions with mid-tier levels of Bloom’s taxonomy, including understanding, applying, and analyzing; skills that are essential for deeper comprehension in K–12 science education.

To more accurately reflect real-world study scenarios, we augment the questions with retrieved lesson content, forming a multimodal retrieval-augmented generation (RAG) pipeline. This approach enables us to evaluate how efficiently MLLMs interpret standalone questions and integrate supplementary multimodal context. 

\vspace{0.3em}
\noindent
\textbf{Contributions.} This paper makes several key contributions to the field. We first benchmark two state-of-the-art multimodal large language models, LLaVA 1.5 and LLaMA 3.2-Vision, on the CK12-QA dataset, evaluating their performance on both diagram-based and text-only questions. This includes a direct comparison of model performance across different input configurations, including standalone questions and retrieval-augmented settings, to assess the impact of contextual information. Furthermore, we conduct a quantitative analysis of the models' ability to align multimodal questions with multimodal context. This work also proposes a retrieval-augmented evaluation framework for textbook question answering, which enables a systematic investigation of context-aware prompting strategies for MLLM-based answer generation. In addition, we identify catastrophic context interference, where adding retrieved context significantly degrades model performance on visual reasoning tasks. This highlights that more context is not always better, underscoring the need for improved context integration in MLLMs. Ultimately, this work advances our understanding of how MLLMs can be leveraged for complex educational reasoning tasks, highlighting both their potential and their current limitations in supporting curriculum-based learning.

\section{Related Work}

\subsection*{Multimodal Question Answering}
Multimodal question answering (QA) has gained increased attention with the advent of vision-language models (VLMs)~\cite{ho2025review} such as BLIP-2~\cite{li2023blip}, MiniGPT-4~\cite{zhu2023minigpt4}, and LLaVA~\cite{liu2023llava}. These models integrate visual encoders with pretrained language models to support image-based reasoning. However, most evaluations focus on datasets like VQAv2~\cite{goyal2017making}, OK-VQA~\cite{marino2019ok}, or COCO~\cite{lin2014microsoft}, which are not designed for the diagram interpretation and complex reasoning needed in educational contexts~\cite{pramanick2024spiqa}. 
Recent developments in educational VQA, such as EduVQA~\cite{xiao2025eduvqa}, apply visual prompting techniques but primarily rely on OK-VQA and do not incorporate retrieval-based context or target fine-grained curriculum-aligned questions. In contrast, our work evaluates pretrained MLLMs, specifically LLaVA and LLaMA 3.2-Vision, on structured educational content, and by analyzes their performance in both standalone and retrieval-augmented settings using lesson-based context.

\subsection*{Knowledge-Augmented MLLMs}
TQA has traditionally been addressed using task-specific architectures that combine retrieval and generation components. Recent advancements have leveraged retrieval-augmented generation (RAG) frameworks and large language models (LLMs) to enhance performance. For instance, the PLRTQA framework~\cite{alawwad2024textbook} integrates RAG with LLM fine-tuning to improve context retrieval and reasoning in out-of-domain scenarios, where relevant knowledge may span multiple textbook lessons. Although PLRTQA does not address the visual aspect of TQA, it demonstrates improved accuracy for textual questions. Its architecture incorporates parameter-efficient fine-tuning and advanced prompt engineering to optimize LLM performance, while context window augmentation helps ensure that retrieved passages provide grounding information, reducing hallucinations and improving answer reliability.
Another promising direction involves integrating knowledge graphs with LLMs~\cite{he2024enhancing}. The TQA-KG method transforms textbook content into structured knowledge graphs, which are then embedded into LLM prompts. This approach enhances reasoning and answer accuracy by enabling the model to identify conceptual relationships, thereby generating more robust and interpretable answers in educational settings.
Earlier efforts, such as ISAAQ~\cite{mavi2022isaaq}, laid the foundation by employing transformer-based language models with both bottom-up and top-down attention mechanisms to address textual and visual comprehension. However, these models were constrained by limited neural capacities and struggled with long-context reasoning and generalization. In contrast, recent knowledge-augmented and LLM-based frameworks represent a substantial advancement, offering improved capabilities to manage the complex, structured, and context-rich demands of textbook question answering.

\section{Method}
\label{sec:method}
\label{sec:method}
Our objective is to evaluate the capacity of modern MLLMs to answer questions grounded in educational content comprising both text and diagrams. To this end, we adopt a multimodal retrieval-augmented generation (RAG) strategy, in which each question is paired with two contextual inputs: a diagram and a paragraph of lesson text retrieved from the CK12-QA corpus.

\subsection*{Multimodal RAG}
We employ a standard retrieval-augmented generation (RAG) approach within a multimodal framework, dynamically extracting relevant paragraphs and diagrams from CK12 lessons for each question. For diagram-based questions, we use the question’s image to retrieve similar diagrams and the accompanying text to retrieve relevant paragraphs. For non-diagram questions, the question text is used to retrieve both pertinent paragraphs and associated diagrams. This strategy ensures that both visual and textual contexts are efficiently leveraged, tailored to the specific question type. 

\paragraph{Similarity-based retrieval.}
Given a query embedding \( f(q) \in \mathbb{R}^d \), derived from the question text or image depending on the modality, and a set of candidate embeddings \( g(c_i) \in \mathbb{R}^d \) representing paragraphs or diagrams, we compute the cosine similarity to rank candidates, as shown in equation~\eqref{eq:cosine}.

\begin{equation}
s(q, c_i) = \frac{\langle f(q), g(c_i) \rangle}{\|f(q)\| \cdot \|g(c_i)\|}
\label{eq:cosine}
\end{equation}

All embeddings are generated using ImageBind~\cite{girdhar2023imagebind}, a cross-modal embedding model that projects inputs from various modalities, such as text and images into a shared embedding space. To store and index paragraph embeddings, we utilize KDB.AI~\cite{kdbai}, a high-performance, scalable vector database optimized for similarity search. During inference, the question embedding is used to query KDB.AI, which efficiently retrieves the top-$k$ context based on their cosine similarity to the query.

The retrieved multimodal context is then combined with the question and its associated diagram (if applicable) to construct a unified prompt, which is input to the MLLM for answer generation. This approach is designed to simulate a realistic educational setting in which students engage with both textual and visual materials from the curriculum.

\subsection*{Multimodal Input Construction}

Each input is composed of several elements to create a comprehensive prompt for the model. It includes a multiple-choice science question, an optional associated diagram such as a labeled process or structure, and a multimodal context consisting of a relevant paragraph and diagrams retrieved from the full set of CK12 lessons using a RAG approach.

\noindent These components are structured into a unified prompt, with the retrieved context presented first, followed by the question and its associated diagram (if applicable). 

In LLaVA-1.5, Prompts alternate between user instructions and assistant responses to simulate a conversational layout. For $N$ retrieved context (paragraphs and images), the model processes $N$ context-image pairs before answering the final question:
\begin{lstlisting}[breaklines=true, basicstyle=\ttfamily\scriptsize]
USER: <retrieved_paragraph_1>
USER: Image#1
ASSISTANT: <caption_response_1>
[...]
USER: Only provide the correct option letter... <question> OPTIONS: <options>
ASSISTANT:
\end{lstlisting}

The LLaMA 3.2-Vision Utilizes a structured chat template with explicit modality tags. Diagrams are embedded using an image token:

\begin{lstlisting}[breaklines=true, basicstyle=\ttfamily\scriptsize]
{"role": "user", "content": [
 {"type": "text", "text": "Context: <retrieved_paragraph>"},
 {"type": "image", "data": <image>},
 {"type": "text", "text": "Image Caption: <caption_1>"},
 ...
 {"type": "text", "text": "Question: <question> Options: <options>"}
]}
\end{lstlisting}

\noindent Both implementations constrain the model to produce letter-only responses (e.g., \texttt{"a"}) to reduce formatting variability. The number of retrieved context paragraphs, denoted as 0p-01, 1p-11, and 3p-3I, corresponds directly to the depth of RAG-based retrieval.

\subsection*{Model Compatibility}
The proposed methodology preserves the original architecture of underlying pretrained multimodal large language models (MLLMs), requiring no structural modifications. Instead, it systematically integrates and further fine-tunes these existing MLLMs within a structured TQA framework. This approach enables task-specific adaptation while maintaining broad applicability and compatibility across diverse model architectures, thereby leveraging both the foundational strengths of pretrained models and the performance gains afforded by targeted fine-tuning.

\begin{table*}[tp]
\centering
\caption{Validation Accuracy (\%) for LLaVA and LLaMA 3.2-Vision under Zero-shot Conditions.}
\label{tab:val_comparison}
\begin{tabular}{lcccccc}
\toprule
\multirow{2}{*}{\textbf{Category}} & \multicolumn{3}{c}{\textbf{LLaVA}} & \multicolumn{3}{c}{\textbf{LLaMA 3.2-Vision}} \\
\cmidrule(lr){2-4} \cmidrule(lr){5-7}
 & \textbf{0p-0I} & \textbf{1p-1I} & \textbf{3p-3I} & \textbf{0p-0I} & \textbf{1p-1I} & \textbf{3p-3I} \\
\midrule
Diagram Qs           & 46.89 & 42.68 & 42.68 & 74.07 & 27.94 & 25.93 \\
T-T/F                & 48.30 & 50.40 & 50.40 & 76.05 & 67.03 & 71.64 \\
T-MC                 & 37.52 & 45.29 & 45.29 & 74.05 & 61.31 & 59.15 \\
All Non-Diagram Qs   & 41.77 & 47.31 & 47.31 & 74.84 & 63.57 & 64.08 \\
All                  & 44.45 & 44.89 & 48.92 & 74.44 & 44.90 & 44.09 \\
\bottomrule
\end{tabular}
\end{table*}

\begin{table*}[tp]
\centering
\caption{Test Accuracy (\%) for LLaVA and LLaMA 3.2-Vision under Zero-shot Conditions.}
\label{tab:test_comparison}
\begin{tabular}{lcccccc}
\toprule
\multirow{2}{*}{\textbf{Category}} & \multicolumn{3}{c}{\textbf{LLaVA}} & \multicolumn{3}{c}{\textbf{LLaMA 3.2-Vision}} \\
\cmidrule(lr){2-4} \cmidrule(lr){5-7}
 & \textbf{0p-0I} & \textbf{1p-1I} & \textbf{3p-3I} & \textbf{0p-0I} & \textbf{1p-1I} & \textbf{3p-3I} \\
\midrule
Diagram Qs           & 40.15 & 35.68 & 37.35 & 74.37 & 20.76 & 15.28 \\
T-T/F                & 51.75 & 53.49 & 53.71 & 73.91 & 66.70 & 68.67 \\
T-MC                 & 39.04 & 48.87 & 53.20 & 74.50 & 60.53 & 57.39 \\
All Non-Diagram Qs   & 43.67 & 50.56 & 53.38 & 74.28 & 62.78 & 61.50 \\
All                  & 41.68 & 42.13 & 44.30 & 74.33 & 38.97 & 35.31 \\
\bottomrule
\end{tabular}
\caption*{\footnotesize 0p-0I, 1p-1I, 3p-3I = number of retrieved paragraphs and images; T-T/F = Textual True/False; T-MC = Textual Multiple Choice; All = overall accuracy across DQ and NDQ.}
\end{table*}

\subsection*{Experimental Setup}
\label{subsec:setup}

We selected LLaVA and LLaMA 3.2-Vision as the MLLMs for our evaluation due to their complementary strengths and alignment with the requirements of our tasks. Both models are open-source, reproducible, and capable of processing vision-language inputs without the constraints of proprietary systems such as OpenAI's GPT-4V~\cite{openai2023gpt4v} or Google's Gemini~\cite{team2023gemini}. LLaMA 3.2-Vision, in particular, represents the state-of-the-art among open-access MLLMs at the time of writing, offering enhanced performance in image understanding and long-context reasoning. By including both models, our study captures a broader perspective on model behavior and generalization in multimodal retrieval and reasoning tasks.

\begin{itemize}
    \item \textbf{LLaVA-1.5}: LLaVA-1.5 is an open-source multimodal large language model that combines CLIP-ViT\cite{radford2021learning} for visual encoding with Vicuna\cite{chiang2023vicuna} for text decoding. This model is instruction-tuned to process both images and text, enabling it to perform tasks such as visual question answering and image-grounded reasoning.
    \item \textbf{LLaMA 3.2-Vision}: LLaMA 3.2-Vision includes pretrained and instruction-tuned generative models with 11 billion parameter. It is capable of processing both text and images to produce text-based outputs. The model is optimized for visual recognition, image reasoning, captioning, and visual question answering.
\end{itemize}

\subsubsection*{Dataset}
We utilize the CK12-QA dataset, a curriculum-aligned benchmark comprising comprising over 26,000 questions derived from the CK-12 Foundation's open educational resources. The dataset includes textual true/false (T-T/F), textual multiple choice (T-MC), and diagram-based (Diagram Qs) items drawn from 1,076 lessons across Life Science, Earth Science, and Physical Science.

\subsubsection*{Input Conditions}

Each model is evaluated under a zero-shot setting using two distinct input configurations to test its reasoning capabilities under different scenarios. The first is a "Basic" condition, where only the question and its associated diagram, if applicable, are provided to the model. The second is a "Multimodal RAG" condition, where a retrieved lesson paragraph is included alongside the question and its diagram to provide additional context.

\subsubsection*{Evaluation Metrics}
We report accuracy rates (\%) for each question category, as well as aggregated scores for all non-diagram questions and overall performance.

\textit{Exact-match (EM) accuracy}

For $N$ question–answer pairs,
\begin{equation}
\text{EM}= \frac{1}{N}\sum_{n=1}^{N}
            \mathbb{1}\!\left[y^{(n)}=\hat y^{(n)}\right],
\label{eq:em}
\end{equation}
where $\hat y^{(n)}$ in equation~\eqref{eq:em} is the model prediction, and $\mathbb{1}[\cdot]$ is the indicator.

\section{Results}
\label{subsec:results}

\subsection{Zero-Shot Performance}

Tables~\ref{tab:val_comparison}--\ref{tab:test_comparison} summarize validation and test accuracy for both models across varying input configurations. LLaMA demonstrates stronger baseline performance without additional context (e.g., 74.50\% vs. 39.04\% on T-MC in the 0p-0I setting--test set). However, LLaVA shows greater relative improvement when context is introduced. In the 3p-3I setting--test set, LLaVA’s T-MC accuracy improves by +14.16\% (from 39.04\% to 53.20\%), narrowing the gap with LLaMA, whose performance declines by –17.11\% (from 74.50\% to 57.39\%) on the test set. This indicates that LLaVA more effectively leverages retrieved textual context, whereas LLaMA’s performance is more sensitive to the presence of distractor paragraphs.

%\subsubsubsection{Diagram Questions (DQ)}  
LLaMA significantly outperforms LLaVA on baseline diagram-based questions, achieving 74.07\% accuracy on the validation set in the 0p-0I setting compared to LLaVA’s 46.89\%. However, with added context, LLaMA’s performance on diagram questions--validation set degrades sharply, dropping to 25.93\% in the 3p-3I setting, while LLaVA remains more stable at 42.89\%. This suggests that LLaVA is more robust to multimodal interference introduced by retrieval-augmented input.

\begin{figure*}[tp]
    \centering
    \includegraphics[width=0.71\textwidth]{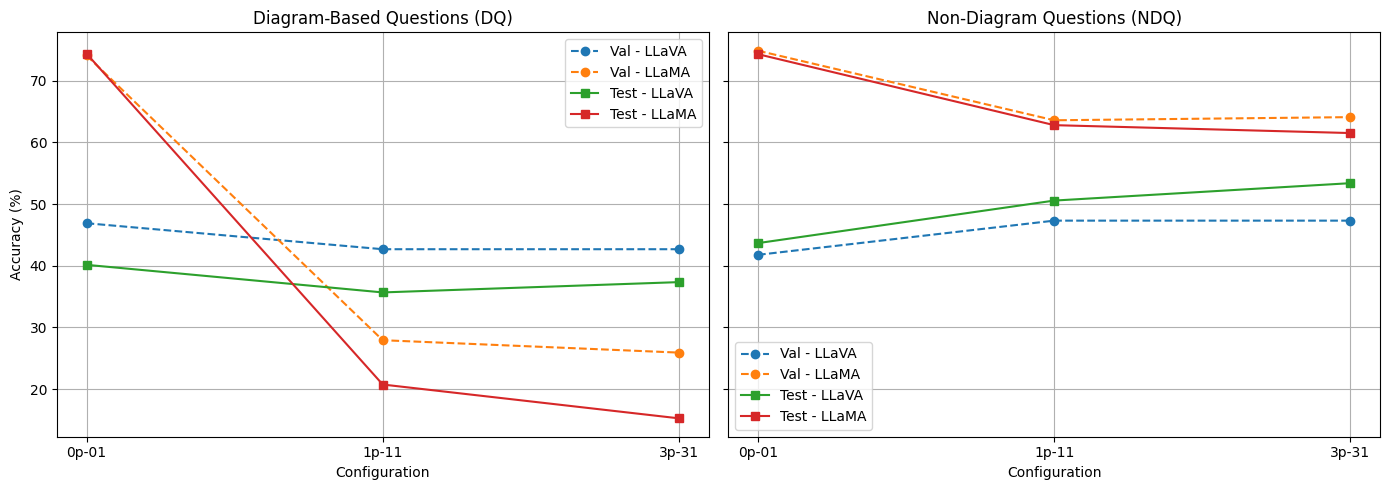} % Replace with your actual file path
    \caption{Accuracy of LLaVA and LLaMA 3.2-Vision across input configurations. The plot compares performance on DQ and NDQ under four input settings, shown for both validation and test sets.}
    \label{fig:model_accuracy}
\end{figure*}

\subsection{Fine-Tuned Model Results}

During fine-tuning, we minimize the autoregressive cross-entropy loss as in equation~\eqref{eq:xent}:
\begin{equation}
\mathcal{L}= -\sum_{t=1}^{T}\log P_\theta\!\bigl(y_t \mid y_{<t},x,I\bigr),
\label{eq:xent}
\end{equation}
where $x$ denotes the retrieved textual context and $I$ represents the diagram image.

We evaluate fine-tuned variants of LLaVA and LLaMA 3.2-Vision under the 3p-3I setting (3 retrieved paragraphs and 3 image) to assess their ability to leverage contextual information following task-specific adaptation.

\begin{table}[H]
\centering
\footnotesize
\caption{Performance of fine-tuned LLaMA 3.2-Vision on CK12-QA under a 3 Paragraph, 3 Image configuration.}
\label{tab:llama_finetuned_comparison}
\begin{tabular}{lcc}
\toprule
\textbf{Metric} & \textbf{Val (3p-3I)} & \textbf{Test (3p-3I)} \\
\midrule
Diagram Qs      & 65.88\% & 63.71\% \\
T-T/F           & 84.67\% & 82.86\% \\
T-MC            & 80.85\% & 79.76\% \\
All Non-Diagram & 82.36\% & 80.89\% \\
All             & 73.72\% & 71.16\% \\
\bottomrule
\end{tabular}
\end{table}

\begin{table}[H]
\centering
\footnotesize
\caption{Performance of fine-tuned LLaVA on CK12-QA under a 3 Paragraph, 3 Image configuration.}
\label{tab:llava_finetuned_comparison}
\begin{tabular}{lcc}
\toprule
\textbf{Metric} & \textbf{Val (3p-3I)} & \textbf{Test (3p-3I)} \\
\midrule
Diagram Qs      & 24.65\% & 25.11\% \\
T-T/F           & 47.29\% & 50.98\% \\
T-MC            & 20.92\% & 19.24\% \\
All Non-Diagram & 31.33\% & 30.81\% \\
All             & 27.78\% & 27.58\% \\
\bottomrule
\end{tabular}
\end{table}

LLaVA shows a marked decline in overall performance following fine-tuning, with accuracy dropping from 48.92\% to 27.78\% on the validation set. Notably, the model exhibits inconsistent behavior across tasks in both test and validation sets, for example, while performance on T-T/F questions had a marginal drop on the test set (from 53.71\% to 50.98\%), accuracy on T-MC questions falls sharply to 19.24\% from 53.20\%, indicating instability in task generalization. Accuracy on diagram-based questions decreased across both validation and test sets, suggesting limited improvement in visual–textual alignment despite fine-tuning.

Fine-tuning outcomes diverge sharply between the two models. As shown in Table~\ref{tab:llama_finetuned_comparison}, LLaMA 3.2-Vision demonstrates substantial improvements over its zero-shot performance, achieving 71.16\% accuracy on the test set, up from 35.31\%. Significant gains are observed across all task types: for example, on the validation set,  T-MC accuracy increases from 59.15\% to 79.76\%, while performance on diagram-based questions rises from 25.93\% to 65.88\%, indicating effective multimodal learning. These results reflect strong cross-modal integration and suggest that fine-tuning successfully mitigates context interference present in zero-shot settings.

In contrast, LLaVA struggles to adapt effectively, as shown in Table~\ref{tab:llava_finetuned_comparison}. Its validation accuracy drops by nearly 50\%, while LLaMA 3.2-Vision achieves 73.72\% overall accuracy on the same split. This disparity highlights potential architectural advantages in LLaMA for multimodal fusion. LLaVA also shows task-specific instability: although T-T/F performance a slight decrease on the test and validation sets, T-MC accuracy declines markedly, pointing to inconsistent task prioritization. By comparison, LLaMA maintains balanced improvements across all tasks, suggesting more robust and curriculum-aware reasoning capabilities. Finally, whereas LLaMA effectively leverages retrieved context and resolves conflicts between textual and diagrammatic inputs, LLaVA appears to either overfit to noisy context or lack sufficient visual grounding, evidenced by its persistently low performance on diagram-based questions.

\textit{Impact of Adding Context}

LLaVA demonstrates notable benefits from retrieved context on text-heavy questions. On the validation set, accuracy for \textit{All non-diagrammatic questions} improves from 41.77\% in the basic setting to 47.31\% under the 3p-3I configuration. A similar trend is observed for T-MC questions on the test set, where accuracy increases from 39.04\% to 53.20\%.
In contrast, LLaMA 3.2-Vision experiences a sharp decline in performance on diagram-based questions when additional context is introduced. On the validation set, accuracy for diagram questions drops from 74.07\% in the basic setting to 25.93\% under 3p-3I, indicating a potential misalignment between textual and visual modalities despite the model’s strong baseline performance.

\textit{Model Comparison}  

A comparison of the models in the zero-shot setting reveals distinct behaviors when handling retrieved context. LLaVA shows a consistent ability to leverage this context, with its overall validation accuracy increasing from 44.45\% in the basic setting to 48.92\% when context is added, highlighting its capacity to integrate additional textual information for enhanced reasoning. In contrast, LLaMA 3.2-Vision, which achieves a much stronger baseline performance on diagram-based questions with 74.07\% accuracy, experiences a sharp performance drop to 25.93\% in the same setting when contextual paragraphs are introduced. This indicates a significant difficulty in fusing retrieved text with its visual reasoning.

\subsection{Catastrophic Context Interference: A Deeper Analysis}

The analysis of these results reveals two primary findings that are central. First, we identify a phenomenon we term "catastrophic context interference," particularly evident in the zero-shot performance of the more capable Llama 3.2 Vision model. While this model demonstrated strong baseline visual reasoning on diagram-based questions (DQs) without any external context (achieving 74.37\% test accuracy), its performance plummeted dramatically to just 15.28\% when a standard MM-RAG pipeline provided three relevant paragraphs and three images as context. This stark degradation highlights a critical failure mode: the naive application of a standard RAG pipeline, intended to provide helpful context, can be actively harmful to the model's reasoning process for certain tasks. The model appears unable to reconcile or filter the retrieved textual information, allowing it to interfere with and override its otherwise strong visual grounding. This finding serves as a crucial cautionary insight for practitioners, demonstrating that more context is not always better and that the interaction between MLLMs and retrieved information is far from a solved problem.

It is important to note that our standard RAG pipeline, while robust, may be providing context that is topically relevant but not optimally salient for answering the specific question. This raises a key question: is the failure in the MLLM's ability to ignore distracting information, or in the retriever's ability to provide only essential information? A crucial direction for future work is to test a "smarter" RAG that uses a re-ranking mechanism to more aggressively filter the context. This would help isolate the variable and more definitively test the MLLM's context integration capabilities.

Second, the study reveals a stark divergence in fine-tuning adaptability between the two models. Llama 3.2 Vision, despite its zero-shot struggles with RAG context, showed a remarkable ability to learn from it during fine-tuning. Its overall test accuracy with RAG context improved from 35.31\% (zero-shot) to 71.16\% after fine-tuning, with performance on DQs recovering substantially. This suggests that its architecture possesses the latent capacity to learn the complex task of integrating and weighing information from multiple multimodal sources when provided with task-specific training data. In contrast, LLaVA 1.5, which showed more modest but stable zero-shot gains from RAG on text-only questions, exhibited a significant performance decline after fine-tuning. This suggests that its shallower modality fusion architecture may be more prone to overfitting on textual cues in the training data, leading to a loss of generalization and persistently poor performance on diagram-centric tasks. This goes beyond a simple "model A is better than model B" conclusion, pointing instead to fundamental differences in architectural depth and capacity for multimodal learning.

\subsection{Qualitative Error Analysis}
To deepen our understanding of why catastrophic context interference occurs, a qualitative analysis of Llama 3.2 Vision's errors in the zero-shot MM-RAG setting was conducted. This analysis revealed several patterns. In many cases, the retrieved text paragraphs, while semantically related to the question's topic, were not directly relevant to answering the specific query posed about the diagram. For example, a question might ask to identify a specific part of the cell cycle in a diagram, while the retrieved text discusses the overall purpose of cell division in broad terms. The model would often seize upon keywords in the distracting text, leading it to an incorrect answer, even when the visual evidence in the diagram was clear.

Second, a common failure mode occurred when the retrieved text described a general scientific principle, but the diagram illustrated a specific instance or exception. The model frequently prioritized the general textual statement over the specific visual evidence, failing to correctly apply the general rule to the specific case shown in the diagram.

Additionally, While the RAG pipeline generally retrieved factually correct information, there were instances where the retrieved text might contain subtle nuances or a slightly different emphasis that conflicted with the simplified representation in the diagram, leading to model confusion.
These examples illustrate that the interference is often not due to factually incorrect retrieved information, but rather due to the MLLM's inability in a zero-shot setting to robustly prioritize sources of information (e.g., trust the specific diagram over the general text) and to identify the most salient piece of information from a noisy, albeit topically relevant, context.

\subsection{Note on Retrieval Quality}
The analysis presented assumes that the underlying MM-RAG pipeline, using ImageBind and KDB.AI for retrieval, provides context that is semantically relevant. A brief inspection of retrieval results confirms that the pipeline consistently retrieves paragraphs and diagrams that are topically related to the query. For instance, questions about the Earth's core retrieve documents about geology and the Earth's layers. Therefore, the observed "catastrophic context interference" is not merely an artifact of poor retrieval but a genuine phenomenon of MLLM reasoning when presented with multiple, sometimes conflicting or distracting, pieces of relevant information. The core issue lies in the integration and utility of the retrieved context, not its topical relevance alone.

\subsection{Discussion}
\label{sec:discussion}

The divergence in post-fine-tuning performance between LLaVA and LLaMA 3.2-Vision offers key insights into the challenges and future directions of multimodal adaptation. While LLaVA exhibits a reasonable zero-shot reasoning capabilities, it fails to generalize effectively after fine-tuning, resulting in a substantial drop in overall accuracy. In contrast, LLaMA 3.2-Vision not only preserves but significantly enhances its performance post-fine-tuning, achieving high results across multiple task categories. This disparity underscores fundamental architectural and training differences between the models.

LLaVA integrates vision and language through a relatively shallow modality fusion architecture, utilizing a frozen CLIP-ViT encoder and a linear projection into Vicuna’s language embedding space. This limited cross-modal interaction constrains the model’s ability to perform deep reasoning, particularly under fine-tuning conditions where it must reconcile potentially conflicting visual and textual signals. Additionally, the use of a frozen visual encoder restricts the model’s capacity to adapt to task-specific visual nuances, leading to brittle and poorly aligned visual–textual representations. In contrast, LLaMA 3.2-Vision benefits from end-to-end multimodal training and the incorporation of deeper cross-attention layers, enabling it to dynamically align and integrate information across modalities more effectively.

Performance under zero-shot settings, as illustrated in Fig.~\ref{fig:model_accuracy}, offers additional insights. LLaMA 3.2-Vision exhibits a sharp decline in diagram-based question accuracy on the validation set, dropping from 74.07\% to 25.93\% when retrieved context is introduced. This degradation suggests a limitation in the model’s ability to filter or reconcile contradictions introduced by noisy or irrelevant textual inputs. In contrast, LLaVA, although generally less proficient in visual reasoning, demonstrates modest improvements on non-diagram questions as contextual information increases. This pattern reinforces the observation that LLaVA’s architecture is more text-centric, favoring textual integration over visual grounding.

\begin{figure}[tb]
  \centering
  \includegraphics[width=0.9\linewidth]{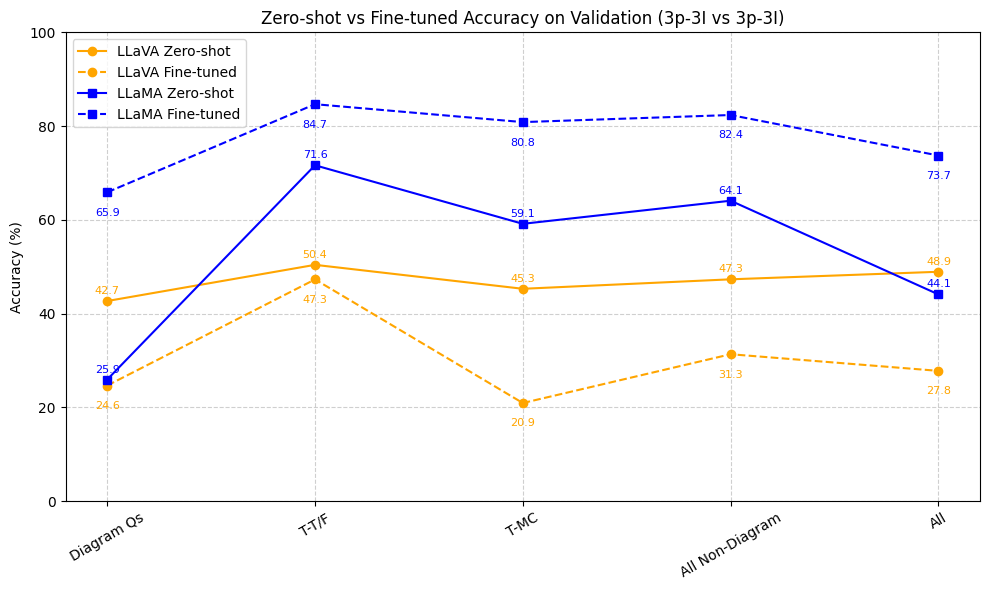}
  \caption{Comparison of zero-shot (3p-3I) and fine-tuned (3p-3I) accuracy on the validation set for LLaVA and LLaMA 3.2-Vision across different question types.}
  \label{fig:val_comparison}
\end{figure}

\begin{figure}[tb]
  \centering
  \includegraphics[width=0.9\linewidth]{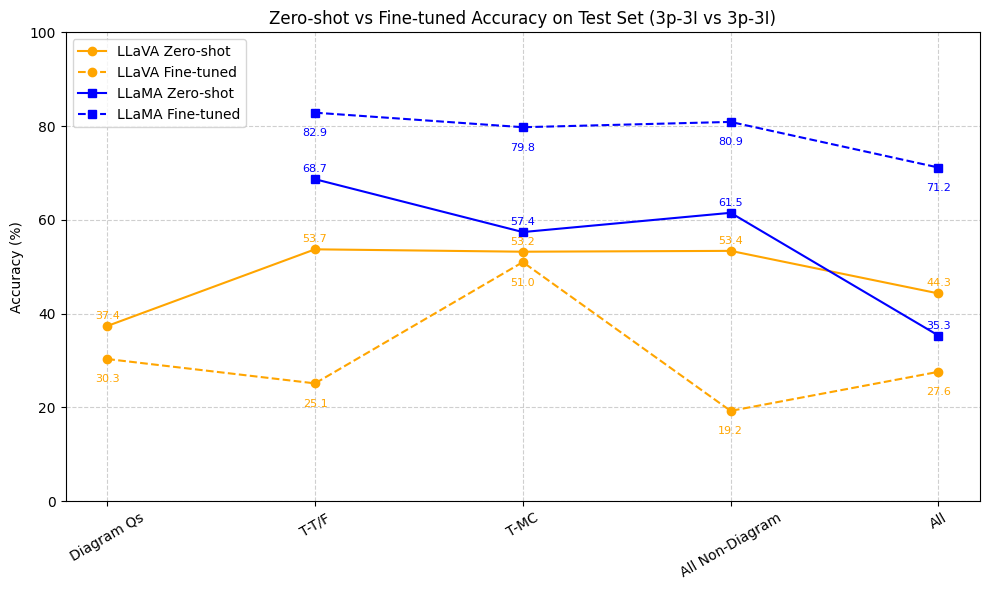}
  \caption{Comparison of zero-shot (3p-3I) and fine-tuned (3p-3I) accuracy on the test set for LLaVA and LLaMA 3.2-Vision across different question types.}
  \label{fig:test_comparison}
\end{figure}

The accuracy of LLaVA and LLaMA 3.2-Vision under zero-shot (3p-3I) and fine-tuned (3p-3I) conditions across different question types is shown in Fig.~\ref {fig:val_comparison} and Fig.~\ref {fig:test_comparison}. The line plots clearly illustrate divergent trends: LLaMA 3.2-Vision achieves substantial improvements following fine-tuning, particularly on both textual and diagram-based questions, while LLaVA exhibits inconsistent gains and notable declines in certain categories. These visualizations underscore the models’ differing capacities to adapt to retrieved multimodal context through fine-tuning on the CK12-QA dataset.

Fine-tuning LLaVA appears to intensify its reliance on textual signals, potentially due to overfitting on a narrowly scoped training set. This results in pronounced task imbalance: although accuracy on T-T/F questions slightly drops, performance on textual T-MC questions deteriorates sharply, and DQ accuracy remains consistently low. These outcomes suggest shortcomings in curriculum-aware reasoning and visual grounding. Contributing factors may include noisy or irrelevant retrieved context, misaligned diagram–text pairs, and inadequate regularization or hyperparameter tuning during training.

This discrepancy underscores the importance of context relevance filtering. Despite their strong vision-language capabilities, both models exhibit difficultly handling indiscriminately retrieved paragraphs. To mitigate this, more effective context selection strategies are needed, strategies that account not only for textual similarity but also for visual, textual alignment. Incorporating cross-modal attention mechanisms may offer a promising solution, enabling models to dynamically evaluate and prioritize context passages based on their relevance to both the question and associated visual input.

Additionally, the analysis reveals a critical gap in modality weighting. LLaVA tends to overemphasize textual cues, whereas LLaMA 3.2-Vision often underutilizes them. Future systems should incorporate adaptive attention mechanisms capable of dynamically balancing visual and textual contributions on a per-instance basis. Such re-weighting could help resolve conflicts between modalities, particularly in cases where one modality carries more informative cues than the other.

Another significant limitation lies in the failure of current retrieval-augmented generation (RAG) approaches to respect curriculum-aware retrieval constraints. Educational datasets such as CK12 exhibit hierarchical knowledge structures, where certain questions depend on definitions, while others require interpretation of annotated diagrams. A curriculum-structured retrieval mechanism, one that accounts for task type and educational progression, could better ensure that retrieved content supports rather than hinders multimodal reasoning.

In summary, the shortcomings observed in LLaVA and the success of LLaMA 3.2-Vision underscore the importance of architectural depth, adaptive training protocols, and task-aligned retrieval strategies for effective multimodal TQA. Addressing these gaps is essential for developing the next generation of educational AI systems grounded in real-world instructional materials.

\section{Conclusion}
\label{sec:conclusion}

This work presents a comprehensive evaluation of two state-of-the-art MLLMs, LLaVA and LLaMA 3.2-Vision, on the CK12-QA dataset, highlighting both their potential and current limitations in educational question answering. Our findings reveal key trade-offs in multimodal integration: while retrieval-augmented context improves performance on text-based questions, it often leads to reduced accuracy on diagram-based questions. For instance, LLaMA 3.2-Vision demonstrates strong visual reasoning in the baseline condition but experiences a significant drop in performance when textual context is introduced. In contrast, LLaVA benefits more from retrieved textual information but exhibits limited gains from visual inputs, underscoring challenges in modality prioritization. These results point to the need for more refined strategies to align and leverage multimodal context effectively. Future work should explore improved retrieval mechanisms that incorporate conceptual dependencies and curricular prerequisites to enhance the relevance and coherence of retrieved materials. Additionally, equipping models with dynamic modality weighting mechanisms could allow them to adaptively balance visual and textual inputs based on question requirements. Finally, instructional fine-tuning on multimodal educational datasets may improve generalization and reduce hallucinations, advancing the reliability and pedagogical effectiveness of AI systems in real-world learning environments.

\bibliographystyle{IEEEtran}
\bibliography{ref}

\begin{thebibliography}{10}
\providecommand{\url}[1]{#1}
\csname url@samestyle\endcsname
\providecommand{\newblock}{\relax}
\providecommand{\bibinfo}[2]{#2}
\providecommand{\BIBentrySTDinterwordspacing}{\spaceskip=0pt\relax}
\providecommand{\BIBentryALTinterwordstretchfactor}{4}
\providecommand{\BIBentryALTinterwordspacing}{\spaceskip=\fontdimen2\font plus
\BIBentryALTinterwordstretchfactor\fontdimen3\font minus \fontdimen4\font\relax}
\providecommand{\BIBforeignlanguage}[2]{{%
\expandafter\ifx\csname l@#1\endcsname\relax
\typeout{** WARNING: IEEEtran.bst: No hyphenation pattern has been}%
\typeout{** loaded for the language `#1'. Using the pattern for}%
\typeout{** the default language instead.}%
\else
\language=\csname l@#1\endcsname
\fi
#2}}
\providecommand{\BIBdecl}{\relax}
\BIBdecl

\bibitem{liu2023llava}
H.~Liu, C.~Li, Q.~Wu, and Y.~J. Lee, ``Visual instruction tuning,'' 2023.

\bibitem{liu2023improvedllava}
H.~Liu, C.~Li, Y.~Li, and Y.~J. Lee, ``Improved baselines with visual instruction tuning,'' \emph{arXiv:2310.03744}, 2023.

\bibitem{grattafiori2024llama}
A.~Grattafiori, A.~Dubey, A.~Jauhri, A.~Pandey, A.~Kadian, A.~Al-Dahle, A.~Letman, A.~Mathur, A.~Schelten, A.~Vaughan \emph{et~al.}, ``The llama 3 herd of models,'' \emph{arXiv preprint arXiv:2407.21783}, 2024.

\bibitem{bewersdorff2025taking}
A.~Bewersdorff, C.~Hartmann, M.~Hornberger, K.~Se{\ss}ler, M.~Bannert, E.~Kasneci, G.~Kasneci, X.~Zhai, and C.~Nerdel, ``Taking the next step with generative artificial intelligence: The transformative role of multimodal large language models in science education,'' \emph{Learn. Individ. Differ.}, vol. 118, p. 102601, 2025.

\bibitem{kuchemann2025opportunities}
S.~K{\"u}chemann, K.~E. Avila, Y.~Dinc, C.~Hortmann, N.~Revenga, V.~Ruf, N.~Stausberg, S.~Steinert, F.~Fischer, M.~Fischer \emph{et~al.}, ``On opportunities and challenges of large multimodal foundation models in education,'' \emph{NPJ Sci. Learn.}, vol.~10, no.~1, p.~11, 2025.

\bibitem{kembhavi2017you}
A.~Kembhavi, M.~Seo, D.~Schwenk, J.~Choi, A.~Farhadi, and H.~Hajishirzi, ``Are you smarter than a sixth grader? textbook question answering for multimodal machine comprehension,'' in \emph{Proc. IEEE Conf. Comput. Vis. Pattern Recognit.}, 2017, pp. 4999--5007.

\bibitem{ho2025review}
H.-T. Ho, L.~V. Nguyen, M.-T. Pham, Q.-H. Pham, Q.-D. Tran, D.~N.~M. Huy, and T.-H. Nguyen, ``A review on vision-language-based approaches: Challenges and applications,'' \emph{Comput. Mater. Continua}, vol.~82, no.~2, 2025.

\bibitem{li2023blip}
J.~Li, D.~Li, S.~Savarese, and S.~Hoi, ``Blip-2: Bootstrapping language-image pre-training with frozen image encoders and large language models,'' in \emph{Proc. Int. Conf. Mach. Learn. (ICML)}, 2023, pp. 19\,730--19\,742.

\bibitem{zhu2023minigpt4}
D.~Zhu, J.~Chen, X.~Shen, X.~Li, and M.~Elhoseiny, ``Minigpt-4: Enhancing vision-language understanding with advanced large language models,'' in \emph{Proc. 12th Int. Conf. Learn. Represent. (ICLR)}, 2024.

\bibitem{goyal2017making}
Y.~Goyal, T.~Khot, D.~Summers-Stay, D.~Batra, and D.~Parikh, ``Making the v in {VQA} matter: Elevating the role of image understanding in visual question answering,'' in \emph{Proc. IEEE Conf. Comput. Vis. Pattern Recognit.}, 2017, pp. 6904--6913.

\bibitem{marino2019ok}
K.~Marino, M.~Rastegari, A.~Farhadi, and R.~Mottaghi, ``Ok-{VQA}: A visual question answering benchmark requiring external knowledge,'' in \emph{Proc. IEEE/CVF Conf. Comput. Vis. Pattern Recognit.}, 2019, pp. 3195--3204.

\bibitem{lin2014microsoft}
T.-Y. Lin, M.~Maire, S.~Belongie, J.~Hays, P.~Perona, D.~Ramanan, P.~Doll{\'a}r, and C.~L. Zitnick, ``Microsoft {COCO}: Common objects in context,'' in \emph{Proc. 13th Eur. Conf. Comput. Vis. (ECCV)}.\hskip 1em plus 0.5em minus 0.4em\relax Zurich, Switzerland: Springer, Sep 2014, pp. 740--755, part 13.

\bibitem{pramanick2024spiqa}
S.~Pramanick, R.~Chellappa, and S.~Venugopalan, ``{SPIQA}: A dataset for multimodal question answering on scientific papers,'' \emph{arXiv preprint arXiv:2407.09413}, 2024.

\bibitem{xiao2025eduvqa}
J.~Xiao and Z.~Zhang, ``Eduvqa: A multimodal visual question answering framework for smart education,'' \emph{Alex. Eng. J.}, vol. 122, pp. 615--624, 2025.

\bibitem{alawwad2024textbook}
H.~A. Alawwad, A.~Alhothali, U.~Naseem, A.~Alkhathlan, and A.~Jamal, ``Enhancing textual textbook question answering with large language models and retrieval augmented generation,'' \emph{Pattern Recognit.}, vol. 162, p. 111332, 2025.

\bibitem{he2024enhancing}
M.~He, A.~Zhou, and X.~Shi, ``Enhancing textbook question answering with knowledge graph-augmented large language models,'' in \emph{Proc. 16th Asian Conf. Mach. Learn. (ACML)}, 2024.

\bibitem{mavi2022isaaq}
\BIBentryALTinterwordspacing
J.~M. Gomez-Perez and R.~Ortega, ``{ISAAQ} - mastering textbook questions with pre-trained transformers and bottom-up and top-down attention,'' in \emph{Proc. 2020 Conf. Empir. Methods Nat. Lang. Process. (EMNLP)}, B.~Webber, T.~Cohn, Y.~He, and Y.~Liu, Eds.\hskip 1em plus 0.5em minus 0.4em\relax Online: Association for Computational Linguistics, Nov. 2020, pp. 5469--5479. [Online]. Available: \url{https://aclanthology.org/2020.emnlp-main.441/}
\BIBentrySTDinterwordspacing

\bibitem{girdhar2023imagebind}
R.~Girdhar, A.~El-Nouby, Z.~Liu, M.~Singh, K.~V. Alwala, A.~Joulin, and I.~Misra, ``Imagebind: One embedding space to bind them all,'' in \emph{Proc. IEEE/CVF Conf. Comput. Vis. Pattern Recognit.}, 2023, pp. 15\,180--15\,190.

\bibitem{kdbai}
{KX}, ``{KDB.AI}: The scalable vector database for ai,'' \url{https://kdb.ai/}, 2023, accessed: 2025-05-22.

\bibitem{openai2023gpt4v}
{OpenAI}, ``{GPT}-4v(ision) system card,'' \url{https://openai.com/index/gpt-4v-system-card/}, 2023, accessed: 2025-05-22.

\bibitem{team2023gemini}
G.~Team, R.~Anil, S.~Borgeaud, J.-B. Alayrac, J.~Yu, R.~Soricut, J.~Schalkwyk, A.~M. Dai, A.~Hauth, K.~Millican \emph{et~al.}, ``Gemini: A family of highly capable multimodal models,'' \emph{arXiv preprint arXiv:2312.11805}, 2023.

\bibitem{radford2021learning}
A.~Radford, J.~W. Kim, C.~Hallacy, A.~Ramesh, G.~Goh, S.~Agarwal, G.~Sastry, A.~Askell, P.~Mishkin, J.~Clark \emph{et~al.}, ``Learning transferable visual models from natural language supervision,'' in \emph{Proc. Int. Conf. Mach. Learn. (ICML)}, 2021, pp. 8748--8763.

\bibitem{chiang2023vicuna}
W.-L. Chiang, Z.~Li, Z.~Lin, Y.~Sheng, Z.~Wu, H.~Zhang, L.~Zheng, S.~Zhuang, Y.~Zhuang, J.~E. Gonzalez \emph{et~al.}, ``Vicuna: An open-source chatbot impressing gpt-4 with 90\% chatgpt quality,'' \url{https://vicuna.lmsys.org}, 2023, accessed: Apr. 14, 2023.

\end{thebibliography}

\end{document}